\documentclass[12pt]{article}

\usepackage{amsmath}
\usepackage{amssymb}
\usepackage{amsfonts}
\usepackage{amsthm}
\usepackage{amscd}
\usepackage{mathrsfs}
\usepackage{bm}
\usepackage{enumerate}
\usepackage{tabu} 
\usepackage{graphicx} 
\usepackage{float} 
\usepackage[all,cmtip]{xy} 
\usepackage{tikz} 
\usepackage{color}
\usepackage{xcolor}
\usepackage[round,comma]{natbib} 

\newcommand{\rd}{\mathrm{d}}

\newcommand{\bbR}{\mathbb{R}}

\newcommand{\barb}{\overline{b}}

\newcommand{\barW}{\overline{W}}

\newcommand{\Rd}{\bbR^d}
\newcommand{\Rdd}{\bbR^{d\times d}}

\newcommand{\p}{\partial}
\newcommand{\id}{\mathrm{id}}

\DeclareMathOperator*{\diag}{diag}
\DeclareMathOperator*{\rank}{rank}

\begin{document}
\title{A Flow Model of Neural Networks
\footnote{Most part of this work was submitted to arXiv as two separated notes on 22 August 
\citep{Zhen_ResNetnoteI} and 6 September respectively. But the latter was not announced 
due to technical reasons. This is a combination of the two previous notes.}
}
\author{
Zhen Li
\thanks{Department of Mathematics, HKUST. \textit{Email: lishen03@gmail.com}}
\quad Zuoqiang Shi
\thanks{Yau Mathematical Sciences Center, Tsinghua University. \textit{Email: zqshi@tsinghua.edu.cn}}
}
\maketitle
\graphicspath{{pics/}}

\abstract

Based on a natural connection between ResNet and transport equation or its characteristic equation,
we propose a continuous flow model for both ResNet and plain net. Through this continuous model,
a ResNet can be explicitly constructed as a refinement of a plain net. The flow model provides
an alternative perspective to understand phenomena in deep neural networks, such as 
why it is necessary and sufficient to use 2-layer blocks in ResNets, why deeper is better, 
and why ResNets are even deeper, and so on. It also opens a gate to bring in more tools from
the huge area of differential equations.

\section{Introduction}

Deep neural networks have been proven impressively successful on certain supervised learning tasks
\citep{nature_deeplearning}. It successively maps datasets to a feature space on which
simple output functions (e.g. softmax classifier) are sufficient to achieve high performance.
Although each single layer is only a simple transformation, the composition of many layers
can represent very complicated functions. Guided by this philosophy and supported by 
powerful computers and massive amount of data, deeper and deeper neural networks are invented
\citep{AlexNet, ZFNet, VGGNet, GoogLeNet}. A remarkable event is that \citet{he2016deep} 
set a new record on the ImageNet competition \citep{imagenet} using their ResNets 
with $152$ and $1202$ layers. Going deeper is believed to be helpful. However, the mechanism 
for that and many other mysteries about the `black box' is still under exploration.

\textbf{Our contributions.}
In this short note, we construct flow models of neural networks. Our aim is not restricted to
answering any specific questions about neural networks, but to build a framework which connects
neural networks with differential equations. As a bridge, it could bring in new perspective 
and new methods, which could be applied to understand or solve learning problems.

We observed that a ResNet is the same as a discretization of the characteristic equation 
of a transport equation. Conversely, the transport equation can be regarded as a 
continuous model of the ResNet. In physics, transport equations are models for describing 
dynamics of quantities which are transported by continuous flows. 
Hence we call the continuous model as a flow model. 

As a natural extension, we also construct a flow model for plain net 
(neural network without residual shortcuts). It is built in a different way. 
This is because non-residual maps between layers can not be considered as discretization
of transport velocity field.

The flow models are immediately available to explain some phenomena in neural networks. 
For example, it naturally supports the belief in the power of depth of neural networks. 
It also relates plain nets to ResNets explicitly. 
The connection is used to explain the super depth of ResNets. Besides, it explains
why it's necessary to use 2-layer blocks in ResNets with ReLU activations, and so on.

\textbf{Related works.}
\citet{Zhen_ResNetcontrol} consider to solve supervised and semi-supervised 
learning problems through PDEs on the point cloud of data.
They propose alternative methods for initializing and training ResNets. 
Recently, we noted that we are not the only ones that observed the connection
between neural networks and differential equations. \citet{E2017proposal} proposes
to study ResNet as a dynamical system. Based on that, \citet{EMSA} consider 
training algorithm from optimal control point of view. 
\citet{Chang2017arXiv} presents an empirical study on the training of ResNet as a dynamical system.
However, all these papers focus on ResNets. We haven't seen any paper 
considering plain nets from similar point of view.

The structure of this note is as follows.
In Section \ref{sec_ResNet}, we start with a transport equation and its characteristic equation
and end up with a ResNet. 
In Section \ref{sec_plainnet}, we build a continuous flow model for a plain net, 
which is done for linear map and activation respectively and then glued up.
In Section \ref{sec_rediscretization}, the flow model of plain net is discretized to get a ResNet.
Considering the relationship between neural networks and their flow models,
we have some comments, which are summarized in Section \ref{sec_discussions}.

\section{Residual Networks}
\label{sec_ResNet}
\subsection{Transport Equation}

Consider the following terminal value problem (TVP) for linear transport equation:
\begin{align}                                                       \label{eq_transport}
\left\{
\begin{aligned}
\p_t u + v(t, x) \cdot\nabla u = 0, &\quad x\in \Rd,~ t \in (0, T)\\
u(T, x) = f(x), &\quad x\in \Rd.
\end{aligned}
\right.
\end{align}
Here $v$ is an $\Rd$-valued function, called transport velocity field.
It can be chosen in different ways. We will consider the general form first, 
then a special type:
\begin{align}
v(t, x) = W^{(2)}(t) a\left(W^{(1)}(t) x + b^{(1)}(t)\right) + b^{(2)}(t),
\label{eq_ResNet_velocity}
\end{align}
where $W^{(1)}(t), W^{(2)}(t)\in\Rdd$, $b^{(1)}(t), b^{(2)}(t)\in\Rd$.
The activation $a$ is an $\Rd$-valued nonlinear function, which is Lipschitz continuous. 

\begin{figure}[H]
\centering
\includegraphics[width=0.6\textwidth]{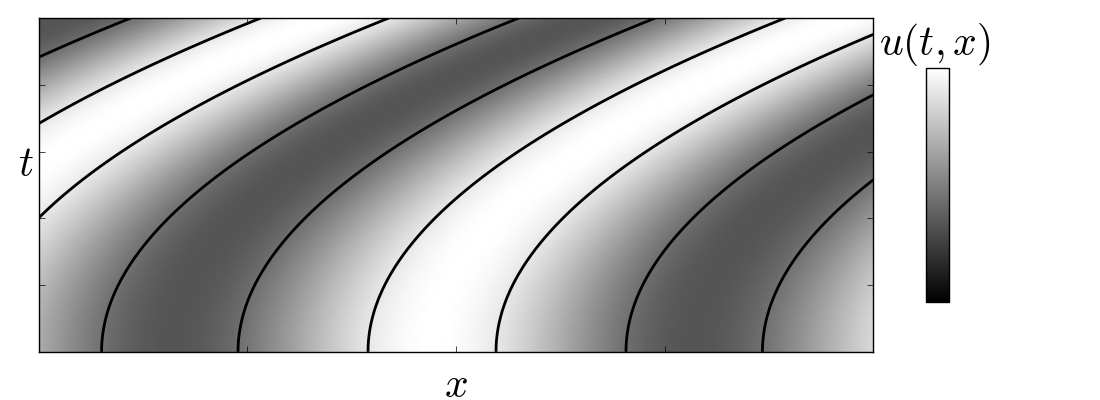}
\caption{Illustration of characteristics. Here $x, u(t, x)\in \bbR$.}\label{fig_char_illu}
\end{figure}

It is well known that the solution of equation \eqref{eq_transport} is transported 
along characteristics, which are defined as solutions of the 
initial value problems (IVP) of the ODE:
\begin{align}                                                       \label{eq_characteristic}
\left\{
\begin{aligned}
&\dot{x} = v(t, x), \quad t \in (0, T)\\
&x(0) = x_0,
\end{aligned}
\right.
\end{align}
where $x_0\in \Rd$. Along the solution curve $x = q(t)$, it is easy to verify that
\begin{align}
\frac{d}{dt} u(t, q(t))
&= (\p_t u(t, x) + \dot{q}(t)\cdot\nabla u(t, x))_{x = q(t)} \\
&= (\p_t u(t, x) + v(t, q(t)) \cdot\nabla u(t, x))_{x = q(t)} = 0.
\end{align}
In the last step we used the transport equation \eqref{eq_transport}.
So $u$ remains unchanged along the curve. 
See Figure \ref{fig_char_illu} for a conceptual illustration. Therefore
\begin{align}                                           \label{eq_backtofuture}
u(0, x_{0}) = u(t, q(t)) = u(T, q(T)) = f(q(T)).
\end{align}
We have solved the transport equation \eqref{eq_transport} by integrating the ODE
\eqref{eq_characteristic}. This is so-called the method of characteristics.

\subsection{Connection with ResNets}

Discretizing the ODE \eqref{eq_characteristic} by Euler's method naturally leads to a ResNet.
In order to make the following approximations reasonable, 
we assume that the change of $v(t, x)$ with $t$ and $x$ is regular enough.
Especially, we assume that the solution of \eqref{eq_transport} and
\eqref{eq_characteristic} exist and are regular enough.

\begin{figure}[H]
\centering
\includegraphics[width=0.6\textwidth]{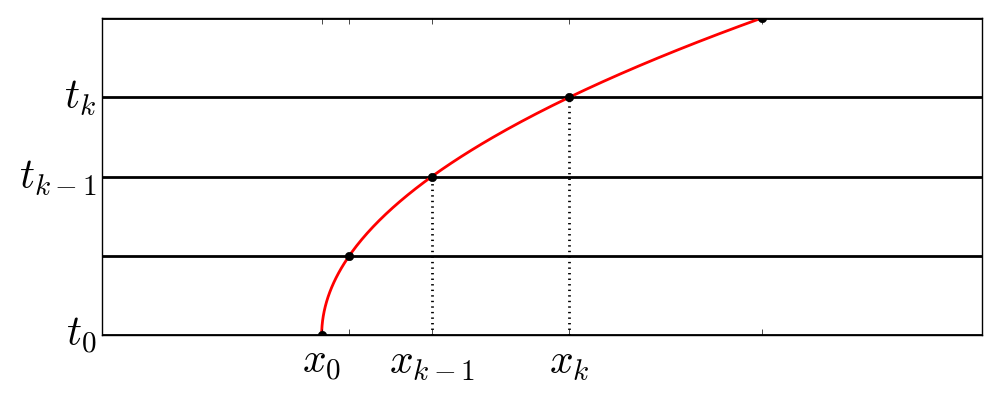}
\caption{Illustration of discretization.}\label{fig_discrete_ODE}
\end{figure}

Let $\{t_k\}_{k = 0}^L$ with $t_0 = 0$ and $t_L = T$ be a partition of $[0, T]\subset\bbR$ 
such that for any $k = 1, \dots, L$, $s_{k} = t_{k}-t_{k-1}$ is small enough. 
Let $x = q(t)$ be a characteristic of the transport equation \eqref{eq_transport},
i.e. the solution of \eqref{eq_characteristic}, and denote $x_k = q(t_k)$. 
Denote $V_{k}(x) = v(t_{k}, x)$ and $u_{k}(x) = u(t_{k}, x)$ for any $x\in\bbR$.
See Figure \ref{fig_discrete_ODE} for a illustration of the discretization.

Near time $t_k$, the ODE \eqref{eq_characteristic} is approximately
\begin{align}                                                       \label{eq_char_approx}
\dot{x} = V_{k}(x_{k}) \approx V_{k}(x_{k-1}).
\end{align}
Use Euler's method to integrate this ODE from $t_{k-1}$ to $t_{k}$, we get
\begin{align}
x_{k}
&\approx x_{k-1} + \int_{t_{k-1}}^{t_{k}} V_{k}(x_{k-1}) \rd t\\
&\approx x_{k-1} + s_{k} V_{k}(x_{k-1})\\
&= (\id + s_{k} V_{k})(x_{k-1}),
\label{eq_ResLayer_general}
\end{align}
where $\id$ is the identity map. Therefore
\begin{align}
x_{L}
&= (\id + s_{L} V_{L})(x_{L-1})\\
&= (\id + s_{L} V_{L})\circ\dots\circ(\id + s_{1} V_{1})(x_{0})
\label{eq_res_curve}
\end{align}
If the terminal value function of $u$ is given as $u_{L} = f$, we might be able to use \eqref{eq_res_curve}
to get the initial value $u_{0}$ at any $x_{0}$. According to \eqref{eq_backtofuture},
\begin{align}
u_{0}(x_{0}) = u_{L}(x_{L})
&= f\circ (\id + s_{L} V_{L})\circ\dots\circ(\id + s_{1} V_{1})(x_{0})
\label{eq_transport_solution}
\end{align}

\begin{figure}[H]
\centering
\includegraphics[width=0.6\textwidth]{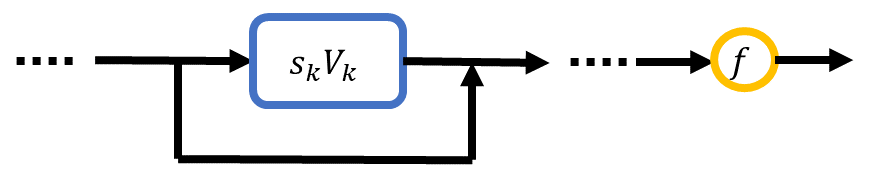}
\caption{Basic structure of a general ResNet. Notice that $\{s_{k} V_{k}\}_{k = 1}^{L}$ 
are generally nonlinear functions of the input.}\label{fig_ResNet}
\end{figure}

The discrete solution \eqref{eq_transport_solution} of the terminal value problem 
of transport equation \eqref{eq_transport} is valid for any $x_{0}\in\Rd$.
Its basic structure is shown in Figure \ref{fig_ResNet}. This structure reminds us
of the ResNet \citep{he2016deep}, but it is merely a formal one.
In order to see the actual structure, we need to specify the definition of $V_{k}$'s.

\begin{figure}[H]
\centering
\includegraphics[width=0.6\textwidth]{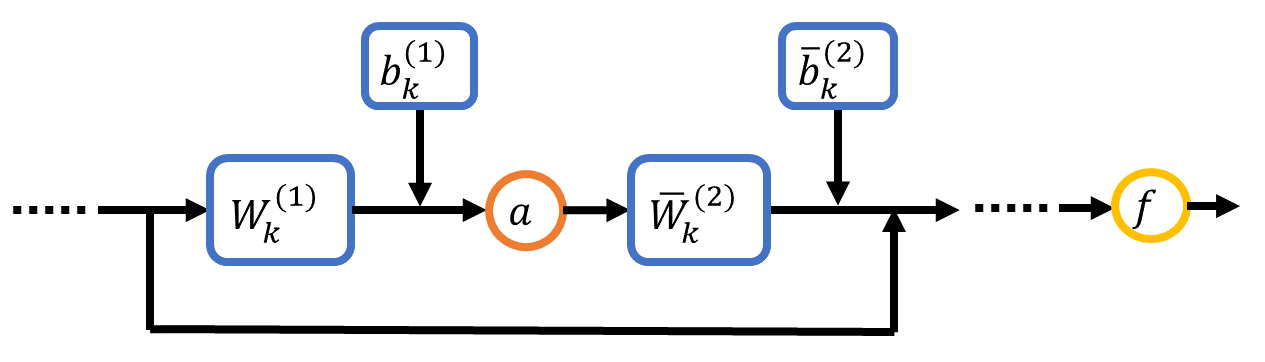}
\caption{Basic structure of the 2-layer ResNet block.}\label{fig_ResNet_2layers}
\end{figure}

\textbf{A Special Type.}
In order to get a ResNet with explicit 2-layer block,
consider the special type of transport velocity field given by \eqref{eq_ResNet_velocity}.
Denote
\begin{align}
W^{(1)}_{k} = W^{(1)}(t_{k}),\quad
&b^{(1)}_{k} = b^{(1)}(t_{k}),\\
W^{(2)}_{k} = W^{(2)}(t_{k}),\quad
&b^{(2)}_{k} = b^{(2)}(t_{k}).\\
\barW^{(2)}_{k} = s_{k} W^{(2)}_{k},\quad
&\barb^{(2)}_{k} = s_{k} b^{(2)}_{k}.
\end{align}
By using the method of characteristics as before, we can get
\begin{align}
x_{k} = x_{k-1} + \barW^{(2)}_{k} ~a\left(W^{(1)}_{k} x_{k-1} + b^{(1)}_{k}\right) + \barb^{(2)}_{k}.
\label{eq_ResNet_2layers}
\end{align}
It generates a $2$-layer ResNet block, which is much more like the original ResNet.
Figure \ref{fig_ResNet_2layers} illustrates its basic structure.

At a first glance, it appears that simply defining the transport velocity as 
\eqref{eq_ResNet_velocity} is not natural. But it is actually reasonable. 
The inner parameters $W^{(1)}$ and $b^{(1)}$ are used to specify the location in the space of data.
It controls where to assigned a velocity vector. If the activation $a$ is non-negative, 
or even restricted to $[0, 1]$, which is often the case, then the outer parameters 
$W^{(2)}$ and $b^{(2)}$ are necessary to adjust the direction and magnitude of the transport velocity.
Both inner parameters and outer parameters are necessary ingredients of the transport velocity field.
Of course, if $a$ is symmetric (such as $\tanh$), the outer parameters are not necessary
for this purpose.

The ResNet obtained here is special.
Firstly, as we can see in \eqref{eq_ResLayer_general} and \eqref{eq_ResNet_2layers}, 
due to the time step $s_{k}$, the residual term can be made sufficiently small 
comparing with the leading term $x_{k}$. This is a necessary condition for the ResNet 
to be modeled by transport equation. 

Secondly, the parameters of the ResNet changes slowly from block to block. 
More specifically, the parameters on the same positions of adjacent ResNet blocks should be close 
to each other, because they are assumed to be discretizations of continuous functions of time.
For example, $W^{(1)}_{k}$ is close to $W^{(1)}_{k-1}$, $W^{(2)}_{k}$ is close to $W^{(2)}_{k-1}$, 
and so on.

\section{Continuous Model of Plain Networks}
\label{sec_plainnet}

We have seen that the method of characteristics for transport equations corresponds to ResNets. 
The key of this connection is the transport velocity field that generates the residual terms between layers.
It's natural to consider similar relationship for a plain net, whose typical layer is
\begin{align}                                                           \label{eq_nonlinear_layer}
x_{k} = a(z_{k}) = a(W_{k} x_{k-1} + b_{k}),
\end{align}
where $a$ is the activation, $W_{k}$ multiplication weight matrix and $b_{k}$ bias vector.
In \eqref{eq_nonlinear_layer}, however, the non-residual term defines a finite (rather than infinitesimal) 
transformation of $x_{k-1}$. It can not be naturally interpreted as velocity,
which makes it difficult to be modeled by transport equation directly.
In this section we will construct a continuous flow for the map \eqref{eq_nonlinear_layer}.
It is done for the linear map and the nonlinear activation respectively.
Later, this flow will be used to construct the ResNet-approximation of the plain net \eqref{eq_nonlinear_layer}.

As a preparation, we define the time scaling function $h(\tau)$. If the flow is only required 
to be continuous in time, then $h(\tau) = \tau$ with $\tau \in [0, 1]$ is sufficient.
Here we require the flow to be smooth, then $h(\tau)$ needs to be nonlinear.
Let $h:~\bbR\rightarrow\bbR$ be a smooth increasing function that satisfies:
\begin{itemize}
\item $h(\tau) = 0$ for $\tau \le 0$,
\item $h(\tau) = 1$ for $\tau \ge 1$,
\item $\dot{h}(\tau) = 0$ for $\tau \notin (0, 1)$.
\end{itemize}
With the above properties of $h(\tau)$, the transport velocity fields of adjacent layers
can be glued up smoothly.
Since we only consider the $k$-th layer in this section, 
let's turn off the suffix $k$ of parameters for simplicity.

\subsection{Linear Map}
\label{sec_approx_linear}
\subsubsection{Approximation by matrix exponentials}

The mainly considered object here is the weight matrix $W$.
Without loss of generality, assume that $x_{k-1}$ and $x_{k}$ has been embedded
into a space of sufficiently high dimension $d$, such that $W$ is a square matrix 
with $\rank(W)<d$. 
If it can be written into an exponential form, we are done.
Unfortunately, this is generally not possible. 
So we consider its full size singular value decomposition
\begin{align}
W = U S V.
\end{align}
Notice that we use $V$ instead of its adjoint $V^{*}$ in the decomposition. 
The requirement that $\rank(W)<d$ is to ensure that $U$ and $V$ can be taken as
proper rotations even if $W$ include mirror reflection on its invariant subspace. 
Since $U$ and $V$ are rotations of finite angles, they can be expressed
as exponential of angular velocity matrices:
\begin{align}
U = \exp \Phi,\quad V = \exp \Psi.
\end{align}
The matrix $S = \diag(e^{\lambda_{1}},\dots,e^{\lambda_{r}}, 0,\dots,0)$ 
is a combination of finite stretch (nonzero diagonals) and projections (zero diagonals). 
But projections can be considered as limits of stretch,
so it can be approximated by a matrix exponential:
\begin{align}
S = \lim_{\beta\to\infty} \exp (\Lambda + \beta\Pi),
\end{align}
where $\Lambda = \diag(\lambda_{1}, ,\dots, \lambda_{r}, 0,\dots,0)$ and
$\Pi = \diag(0,\dots,0, -1,\dots,-1)$ where the last $d-r$ entries are $-1$.
Thus
\begin{align}
W &= \lim_{T\to\infty} \exp(\Phi) \exp(\Lambda + \beta\Pi) \exp(\Psi)\\
&\approx \exp(\Phi) \exp(\Lambda + \beta\Pi) \exp(\Psi)
\end{align}
for large $\beta$. So the map \eqref{eq_nonlinear_layer} can be approximated by
\begin{align}                                                           \label{eq_approx_exp_layer}
z_{k} = \exp(\Phi) \exp(\Lambda + \beta\Pi) \exp(\Psi) x_{k-1} + b.
\end{align}

\subsubsection{Flows of linear maps}
The linear map \eqref{eq_approx_exp_layer} can be approached by a composition of continuous flows.
Denote
\begin{align}
P(\tau) = \exp(h(\tau) \Phi), \quad
Q(\tau) = \exp(h(\tau)(\Lambda + \beta\Pi)),\quad
R(\tau) = \exp(h(\tau) \Psi),
\end{align}
then $P(0) = Q(0) = R(0) = \id$ and $P(1) = U, Q(1)\approx S, R(1) = V$. 
For any $x\in\Rd$, define the translation flow
\begin{align}
\psi(\tau, x) = x + h(\tau) b.
\end{align}
Then $\psi(0, x) = x$ and $\psi(1, x) = x + b$.
So the linear  maps $U, S, V$ and $\psi(1,\cdot)$ can be modeled by continuous flows,
each takes one unit of time. In the following we consider their transport velocity fields.

The rotation flow $R(\tau)$ can also be described by the initial value problem of ODE
\begin{align}
\left\{
\begin{aligned}
&\dot{z}(\tau) = \dot{h}(\tau)\Psi z, &\tau\in[0, 1],\\
&z(0) = x_{k-1},
\end{aligned}
\right.
\end{align}
because its solution is just 
\begin{align}
z = \xi (\tau, x_{k-1}) = R(\tau)x_{k-1}.
\end{align}
It means that the transport velocity field
from $\xi(0, x_{k-1}) = x_{k-1}$ to $V x_{k-1}$ is defined by $\dot{h}(\tau)\Psi z$.

In a similar way, the stretch flow $Q(\tau)$ can also be described by
\begin{align}
\left\{
\begin{aligned}
&\dot{z}(\tau) = \dot{h}(\tau)(\Lambda + \beta\Pi) z, &\tau\in[0, 1],\\
&z(0) = V x_{k-1},
\end{aligned}
\right.
\end{align}
because its solution is just 
\begin{align}
z = \eta (\tau, V x_{k-1}) = Q(\tau)V x_{k-1}. 
\end{align}
It means that the transport velocity field
from $\eta(0, V x_{k-1}) = V x_{k-1}$ to $SV x_{k-1}$ is defined by $\dot{h}(\tau)(\Lambda + \beta\Pi) z$.

In a similar way, the rotation map $P(\tau)$ can also be described by
\begin{align}
\left\{
\begin{aligned}
&\dot{z}(\tau) = \dot{h}(\tau)\Phi z, &\tau\in[0, 1],\\
&z(0) = SV x_{k-1},
\end{aligned}
\right.
\end{align}
because its solution is just 
\begin{align}
z = \zeta (\tau, SV x_{k-1}) = P(\tau)SV x_{k-1}. 
\end{align}
It means that the transport velocity field
from $\zeta(0, SV x_{k-1}) = SV x_{k-1}$ to $W x_{k-1}$ is defined by $\dot{h}(\tau)\Phi z$.

Finally, the translation flow $\psi(\tau,\cdot)$ can also be described by
\begin{align}
\left\{
\begin{aligned}
&\dot{z}(\tau) = \dot{h}(\tau) b, &\tau\in[0, 1],\\
&z(0) = USVx_{k-1} = W x_{k-1},
\end{aligned}
\right.
\end{align}
because its solution is simply 
\begin{align}
z = \psi(\tau, W x_{k-1}).
\end{align}
It means that the transport velocity field 
from $W x_{k-1}$ to $W x_{k-1} + b$ is just $\dot{h}(\tau) b$.

By Euler method, it can be shown that the linear exponential layers $U, S, V$
can all be approximated by several linear ResNet blocks.
\subsection{Activation}
Now let's consider the nonlinear activation $a$.
Assume that $a$ is non-decreasing, differentiable almost everywhere and Lipschitz. 
From now on, denote
\begin{align}
z_{k} = W x_{k-1} + b,
\end{align}
then we have $x_{k} = a(z_{k})$. For any $Z\in\Rd$ and $\tau \in [0, 1]$, define
\begin{align}                                                                   \label{eq_activation_flow}
\varphi(\tau, Z) = (1 - h(\tau)) Z + h(\tau) a(Z).
\end{align}
Clearly,
\begin{align}
\varphi(0, Z) = Z, &\quad \varphi(1, Z) = a(Z).
\end{align}
So it takes one unit of time to move from $Z$ to $a(Z)$.
Fix any $\tau \in [0, 1)$, the value of $\varphi(\tau, Z)$ is strictly increasing in $Z$, hence invertible.
Denote $z(\tau) = \varphi(\tau, Z)$, hence $Z = \varphi^{-1}(\tau, z(\tau))$.
As $\tau$ goes from $0$ to $1$, $\varphi$ is a flow that continuously moves $z_{k}$ to $x_{k}$.
The transport velocity field is given by
\begin{align}
v^{a}(\tau, z(\tau)) 
&= \p_{\tau}\varphi(\tau, Z)\\
&= \dot{h}(\tau) (a(Z) - Z)\\
&= \dot{h}(\tau) \left(a\left(\varphi^{-1}(\tau, z(\tau))\right) - \varphi^{-1}(\tau, z(\tau))\right),
\end{align}
or simply
\begin{align}                                                                       \label{eq_activation_velocity}
v^{a}(\tau, z) 
= \dot{h}(\tau) \left(a\left(\varphi^{-1}(\tau, z)\right) - \varphi^{-1}(\tau, z)\right).
\end{align}
Thus $\varphi(\tau, z_{k})$ is the solution to the initial value problem
\begin{align}                                                                       \label{eq_activation_char}
\left\{
\begin{aligned}
&\dot{z} = v^{a}(\tau, z),\\
&z(0) = z_{k},
\end{aligned}
\right.
\end{align}
and $x_{k} = \varphi(1, z_{k})$.

\textbf{Example.}
Before moving on, let's look at an example of activation flow $\varphi(\tau, Z)$.
Let $a$ be ReLU. For any $Z\in\bbR$,
\begin{align}
a(Z) = \max(Z, 0) =
\left\{
\begin{aligned}
&Z, &\quad Z \ge 0,\\
&0, &\quad Z < 0.
\end{aligned}
\right.
\end{align}
By definition \eqref{eq_activation_flow}, the activation flow is
\begin{align}
\varphi(\tau, Z) = \max(Z, (1-h(\tau)) Z) =
\left\{
\begin{aligned}
&Z, &\quad Z \ge 0,\\
&(1-h(\tau)) Z, &\quad Z < 0.
\end{aligned}
\right.
\end{align}
Notice that for any $\tau \in [0, 1)$, it is a leaky ReLU.
If $z = \varphi(\tau, Z)$, then
\begin{align}
\varphi^{-1}(\tau, z) = \min\left(z, \frac{z}{1-h(\tau)}\right) =
\left\{
\begin{aligned}
&z, &\quad z \ge 0,\\
&\frac{z}{1-h(\tau)}, &\quad z < 0.
\end{aligned}
\right.
\end{align}
Hence the transport velocity field is
\begin{align}
v^{a}(\tau, z) 
&= \frac{\dot{h}(\tau)}{1-h(\tau)}(a(z) - z)
= \left\{
\begin{aligned}
&0, &\quad z \ge 0,\\
&-\frac{\dot{h}(\tau)}{1-h(\tau)}z, &\quad z < 0,
\end{aligned}
\right.\\
&= a\left( \frac{\dot{h}(\tau)}{h(\tau) - 1} z \right).
\label{eq_activation_velocity_ReLU}
\end{align}

\subsection{Gluing Up}
In summary, the map of nonlinear plain layer \eqref{eq_nonlinear_layer} can be
modeled successively by the flows $\xi$, $\eta$, $\zeta$, $\psi$, $\varphi$.
So it takes $4$ units of time to move from $x_{k-1}$ to $z_{k} = W x_{k-1} + b$,
then takes one unit of time to move from $z_{k}$ to $x_{k} = a(z_{k})$.
For technical completeness, let's glue this flows together. 
For any $x\in\Rd$, define
\begin{align}
\theta^{k}(\tau, x) =
\left\{
\begin{aligned}
\xi(\tau, x&), &\quad &\tau \in [0, 1),\\
\eta(\tau-1, \xi(1, x&)), &\quad &\tau \in [1, 2),\\
\zeta(\tau-2, \eta(1, \xi(1, x&))), &\quad &\tau \in [2, 3),\\
\psi(\tau-3, \zeta(1, \eta(1, \xi(1, x&)))), &\quad &\tau \in [3, 4),\\
\varphi(\tau-4, \psi(1, \zeta(1, \eta(1, \xi(1, x&))))), &\quad &\tau \in [4, 5).
\end{aligned}
\right.
\label{eq_layer_flow}
\end{align}
For convenience, the above sequentially glued flow \eqref{eq_layer_flow}
is called the layer flow of the $k$-th layer.

The layer flow \eqref{eq_layer_flow} can also be described by the ODE
\begin{align}
\dot{z} = v^{k}(\tau, z) =
\left\{
\begin{aligned}
&\dot{h}(\tau) \Psi z, &\quad &\tau \in [0, 1),\\
&\dot{h}(\tau - 1) (\Lambda + \beta\Pi) z, &\quad &\tau \in [1, 2),\\
&\dot{h}(\tau - 2) \Phi z, &\quad &\tau \in [2, 3),\\
&\dot{h}(\tau - 3) b, &\quad &\tau \in [3, 4),\\
&\dot{h}(\tau - 4) \left(a\left(\varphi^{-1}(\tau - 4, z)\right) - \varphi^{-1}(\tau - 4, z)\right), 
            &\quad &\tau \in [4, 5),
\end{aligned}
\right.
\label{eq_transport_velocity}
\end{align}
with initial condition $z(0) = x_{k-1}$. Then $x_{k} = z(5)$.
Notice that at the $\tau = 0, \dots, 5$, the velocity vanishes.

Notice that the above sequentially gluing procedure is only one of possible ways
to construct a continuous flow for \eqref{eq_nonlinear_layer}. There are infinitely
many flows that produce the same nonlinear map \eqref{eq_nonlinear_layer}, 
although most of them do not have such explicit formulation.

Previously, we construct a transport velocity field for a typical single layer of plain net.
Let's construct the velocity field for the whole network.
Consider the terminal value problem of the linear transport equation \eqref{eq_transport}.
Now the transport velocity field $v$ is defined by gluing up \eqref{eq_transport_velocity}
of different layers.
The detail is as follows. Let $\{t_{k}\}_{k = 0}^L$ with $t_{0} = 0$ and $t_{L} = T$ 
be a uniform partition of $[0, T]$ such that $s_{k} = t_{k} - t_{k-1}$ be small enough. 
Then for $k = 1, \dots, L$,
\begin{align}
v(t, x) := v^{k}\left( \frac{5}{s_{k}}(t - t_{k-1}), x \right), \quad& t\in [t_{k-1}, t_{k}).
\end{align}
Notice that the time is scaled such that $s$ units of time here is equivalent 
to $5$ units of time in \eqref{eq_transport_velocity}.
Notice that for any $k = 0, \dots, L$, the transport velocity field $v(t_{k}, x) = 0$.
It means that $v(t, x)$ is smooth in $t$.
Thus we have seen that the transport equation is a continuous model for the plain net.
Given any plain net, we can construct a transport equation using its parameters and activations.

\section{Re-Discretization as ResNet}
\label{sec_rediscretization}

In Section \ref{sec_ResNet}, we have shown that ResNets can be modeled by
continuous flows. In Section \ref{sec_plainnet}, we have shown that plain nets
can also be modeled by continuous flows. 
It's natural to consider the connection of the two types of neural networks
through their continuous models. In this section,
we show that by re-discretizing the flow model obtained from the plain net, 
we can get a ResNet, which is an approximation of the plain net.
More specifically, each layer of the plain net is approximated by several ResNet blocks.

\subsection{Linear map}
We have two options for the linear map
\begin{align}                                                           \label{eq_linear_layer}
z_{k} = W x_{k-1} + b
\end{align}
One option is to leave it as a whole map. The other option is to discretize its 
continuous model in the same way as we did in Section \ref{sec_ResNet}.
For the second option, one only needs to applying Euler's method to the ODEs 
in \eqref{eq_transport_velocity} which corresponds to the linear map. 
Let's discretize the first equation in \eqref{eq_transport_velocity} as an example.
Let $\{\tau_{r}\}_{r = 0}^l$ with $\tau_{0} = 0$ and $\tau_{l} = 1$ be a uniform partition of $[0, 1]$,
such that $\alpha = \tau_{r+1} - \tau_{r}$ is small enough. 
Denote $y_{r} = z(\tau_{r})$, then $y_{0} = x_{k-1}$. By Euler's method, we have
\begin{align}
y_{r+1} 
&= y_{r} + \alpha \dot{h}(\tau_{r}) \Psi y_{r},
\end{align}
which is a linear 1-layer ResNet block. Repeat this iteration for $l$ times, we have
\begin{align}
y_{l}                                                           \label{eq_linear_map_by_ResNet}
&= \left( \id + \alpha \dot{h}(\tau_{l}) \Psi \right) \circ \cdots \circ
    \left( \id + \alpha \dot{h}(\tau_{1}) \Psi \right) y_{0}.
\end{align}
We can apply the same procedure to the second, third and fourth equation in 
\eqref{eq_transport_velocity}.
The discretization of these equations are very similar, hence are omitted here.

\subsection{Activation}
In the following, let's focus on the nonlinear part.
The activation flow is solved from \eqref{eq_activation_char} in the following way.
Recall that it takes one unit of time to move from $z_{k}$ to $x_{k} = a(z_{k})$.
For clarity in notations, we still use $[0, 1]$ as the range of time $\tau$.
Let $\{\tau_{r}\}_{r = 0}^l$ with $\tau_{0} = 0$ and $\tau_{l} = 1$ be a uniform partition of $[0, 1]$,
such that $\alpha = \tau_{r+1} - \tau_{r}$ is small enough.
Denote $y_{r} = z(\tau_{r})$ and $v^{a}_{r} = v^{a}(\tau_{r}, y_{r})$.
Then $z_{k} = y_{0}$ and $x_{k} = a(z_{k}) = y_{l}$.
Solve \eqref{eq_activation_char} by Euler method iteratively, we have
\begin{align}
y_{r+1} 
&= y_{r} + \alpha v^{a}_{r}\\
&= y_{r} + \alpha \dot{h}(\tau_{r}) 
\left(a\left(\varphi^{-1}(\tau_{r}, y_{r})\right) 
- \varphi^{-1}(\tau_{r}, y_{r})\right)
\label{eq_activation_iteration}
\end{align}
To see the basic structure of ResNet, let's make \eqref{eq_activation_iteration} explicit.

\textbf{Example.}
For ReLU activation $a$, it is straightforward. According to \eqref{eq_activation_velocity_ReLU},
\begin{align}
v^{a}_{r} 
= v^{a}(\tau_{r}, y_{r})
= a\left( \frac{\dot{h}(\tau_{r})}{h(\tau_{r}) - 1} y_{r} \right).
\end{align}
Therefore,
\begin{align}
y_{r+1} 
&= y_{r} + \alpha a\left( \frac{\dot{h}(\tau_{r})}{h(\tau_{r}) - 1} y_{r} \right)\\
&= y_{r} + a\left( \frac{\alpha \dot{h}(\tau_{r})}{h(\tau_{r}) - 1} y_{r} \right),
\end{align}
which is a 1-layer ResNet block with scalar weight
\begin{align}
W_{r} = \frac{\alpha \dot{h}(\tau_{r})}{h(\tau_{r}) - 1} \id
\end{align}
and $b_{r} = 0$. 
Thus for ReLU activation, the approximation of plain nets by ResNets is quite trivial.

\begin{figure}[H]
\centering
\includegraphics[width=0.7\textwidth]{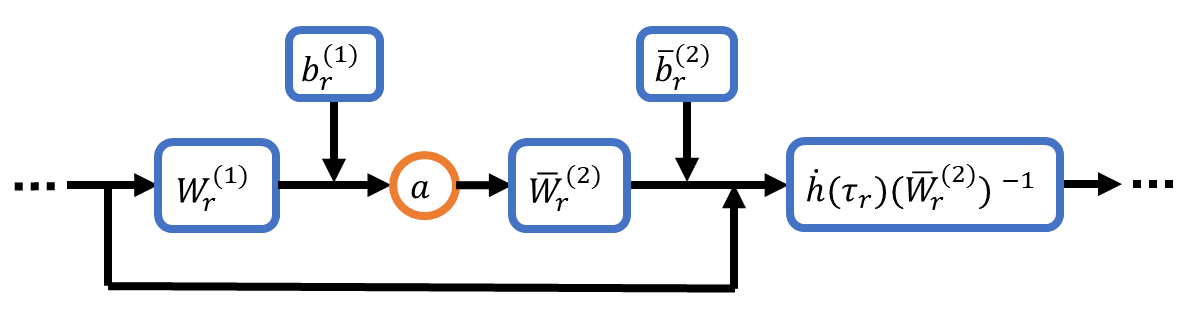}
\caption{One iteration of discretized activation flow. 
Several iterations are needed to approximate the activation in one layer of a plain net.
(Best see in color.)}\label{fig_ResNet_activation}
\end{figure}

If $\varphi^{-1}(\tau, \cdot)$ has no explicit expression or is nonlinear, 
we may consider its linearization at $\tau_{r}$ and near $y_{r}$. 
According to the definition of $\varphi$ \eqref{eq_activation_flow}, 
the Jacobian of $\varphi(\tau, \cdot)$ at any $Z$ is
\begin{align}
J(\tau, Z) 
&= \frac{\p \varphi(\tau, Z)}{\p Z} \\
&= (1-h(\tau)) \id + h(\tau) \diag(a'(Z))\\
&= \diag\left((1-h(\tau)) + h(\tau) a'(Z)\right),
\end{align}
whose inverse is the inverse Jacobian in terms of $Z$:
\begin{align}
J^{-1}(\tau, Z) 
&= \diag\left(\frac{1}{(1-h(\tau)) + h(\tau) a'(Z)}\right).
\end{align}
Notice that $Z$ is a vector and the fraction is entry-wise.
Since the linearization of inverse is the inverse of linearization,
we first linearize $\varphi(\tau, Z)$ at $y_{0} = z_{k}$, then compute its inverse.
\begin{align}
z = \varphi(\tau, Z)
&\approx \varphi(\tau, z_{k}) + J(\tau, z_{k})(Z-z_{k}),
\end{align}
therefore
\begin{align}
Z = \varphi^{-1}(\tau, z)
&\approx J^{-1}(\tau, z_{k}) z + z_{k} - J^{-1}(\tau, z_{k}) \varphi(\tau, z_{k}).
\end{align}
For simplicity, denote
\begin{align}
W^{(1)}_{r} &= J^{-1}(\tau_{r}, z_{k}),\\
b^{(1)}_{r} &= z_{k} - J^{-1}(\tau_{r}, z_{k}) \varphi(\tau_{r}, z_{k}),
\end{align}
and take $z = y_{r}$, we have
\begin{align}
\varphi^{-1}(\tau_{r}, y_{r}) \approx W^{(1)}_{r} y_{r} + b^{(1)}_{r}.
\end{align}
Then the iteration \eqref{eq_activation_iteration} becomes
\begin{align}
y_{r+1} 
&= y_{r} + \alpha \dot{h}(\tau_{r}) 
\left(a\left(W^{(1)}_{r} y_{r} + b^{(1)}_{r}\right) 
- W^{(1)}_{r} y_{r} - b^{(1)}_{r}\right)\\
&= \left( \id - \alpha \dot{h}(\tau_{r}) W^{(1)}_{r} \right) y_{r}
+ \alpha \dot{h}(\tau_{r}) a\left(W^{(1)}_{r} y_{r} + b^{(1)}_{r}\right)
- \alpha \dot{h}(\tau_{r}) b^{(1)}_{r}.
\end{align}
Let
\begin{align}
\left\{
\begin{aligned}
&W^{(2)}_{r} = \dot{h}(\tau_{r}) \left( \id - \alpha \dot{h}(\tau_{r}) W^{(1)}_{r} \right)^{-1},\\
&b^{(2)}_{r} = - W^{(2)}_{r} b^{(1)}_{r},\quad
\end{aligned}
\right.
\quad
\left\{
\begin{aligned}
&\barW^{(2)}_{r} = \alpha W^{(2)}_{r}, \\
&\barb^{(2)}_{r} = \alpha b^{(2)}_{r}.
\end{aligned}
\right.
\end{align}
Then we get
\begin{align}                                                               \label{eq_activation_by_ResNet}
y_{r+1} = \dot{h}(\tau_{r}) \left(W^{(2)}_{r}\right)^{-1} 
\left( y_{r}
+ \barW^{(2)}_{r} a\left(W^{(1)}_{r} y_{r} + b^{(1)}_{r}\right)
+ \barb^{(2)}_{r} \right),
\end{align}
which is the approximation of \eqref{eq_activation_iteration}.
It contains a $2$-layer ResNet block followed by a non-residual linear map, 
as shown in Figure \ref{fig_ResNet_activation}.
The whole activation flow is composed of several iterations in \eqref{eq_activation_iteration}
or its approximation \eqref{eq_activation_by_ResNet}.

\begin{figure}[H]
\centering
\includegraphics[width=0.7\textwidth]{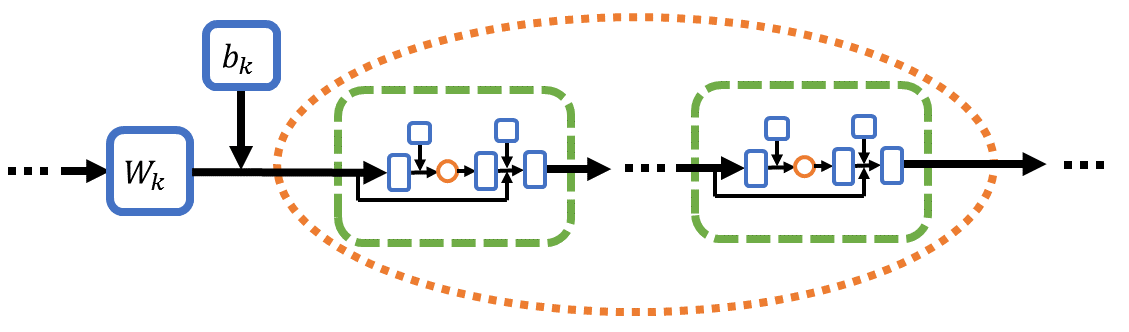}
\caption{A single layer of plain net approximated by the composition of a linear map 
and a multi-layer ResNet.
Each dashed green box represents a structure in Figure \ref{fig_ResNet_activation}.
The dotted orange ellipse approximate the activation in one layer of a plain net.
(Best see in color.)}\label{fig_1layer_by_hybrid_ResNet}
\end{figure}

\begin{figure}[H]
\centering
\includegraphics[width=0.9\textwidth]{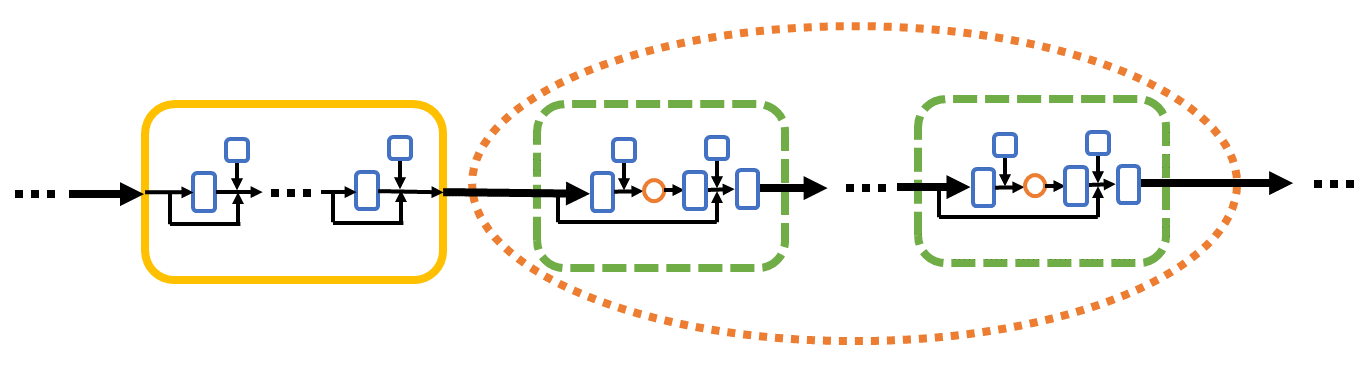}
\caption{Alternative to Figure \ref{fig_1layer_by_hybrid_ResNet}.
The solid yellow box contains a composition of linear 1-layer ResNet blocks, 
which approximates the linear map of the original plain net.
(Best see in color.)}\label{fig_1layer_by_pure_ResNet}
\end{figure}

Together with the linear map \eqref{eq_linear_layer}, the single $k$-th layer of the plain net 
\eqref{eq_nonlinear_layer} is approximated by the composition of linear maps and 2-layer ResNet blocks. 
See Figure \ref{fig_1layer_by_hybrid_ResNet}.
Alternatively, we can also use \eqref{eq_linear_map_by_ResNet} and its successors
instead of the whole linear map \eqref{eq_linear_layer}. See Figure \ref{fig_1layer_by_pure_ResNet}.

Now it may be a little confusing: The multi-layer ResNet still contains several activations
(the small orange circles with solid border line, within each dashed green box 
in Figures \ref{fig_1layer_by_hybrid_ResNet} and \ref{fig_1layer_by_pure_ResNet}). 
Why bother to replace one activation (the orange dotted ellipse) 
by such a multi-layer structure containing more activations? The answer is as follows.
The roles of activations in original plain net and in the new ResNet are different.
In the plain net, the activation causes nonlinear distortion to the map between two layers,
or poses geometric constraint on the layer flow. The effect is significant and immediate.
In the ResNet got above, however, the activation causes nonlinear distortion to 
the transport velocity field, or poses differential constraint on the layer flow.
The effect can become significant only after accumulation.

Another confusing thing is about the continuously change of parameters from layer to layer.
Since the neural networks here are got from discretizing a continuous flow,
it's natural to guess that the parameters of the networks varies slowly from layer to layer.
However we should be careful about this idea. It is generally not true for nonlinear networks.
For the nonlinear plain net \eqref{eq_nonlinear_layer}, we have seen in \eqref{eq_transport_velocity}
that the continuous transport velocity field is NOT simply like
\begin{align}
v(t, x) = a(W(t) x + b(t)).
\end{align}
So the parameters $W_{k}$ and $b_{k}$ in \eqref{eq_nonlinear_layer} themselves 
should not be regarded as discretization of some continuous parameters of a velocity field. 

For the nonlinear ResNet shown in Figure \ref{fig_1layer_by_hybrid_ResNet}, the situation is more subtle.
To approximate the activation in one layer of the original plain net \eqref{eq_nonlinear_layer}
(the dotted orange ellipse), several basic structures (the dashed green boxes in Figure \ref{fig_1layer_by_hybrid_ResNet})
of the ResNet are used.
The structure of these basic structures are the same, the parameters on their corresponding positions
varies slowly. It means that as $r$ changes, $W^{(1)}_{r}$ changes slowly, 
$b^{(1)}_{r}$ changes slowly, and so on. In this sense, the parameters of ResNet changes continuously.
But if we only naively go through the parameters layer by layer, we will not find this continuity.

\section{Discussions}
\label{sec_discussions}

\begin{figure}[H]
\centering
\includegraphics[width=0.6\textwidth]{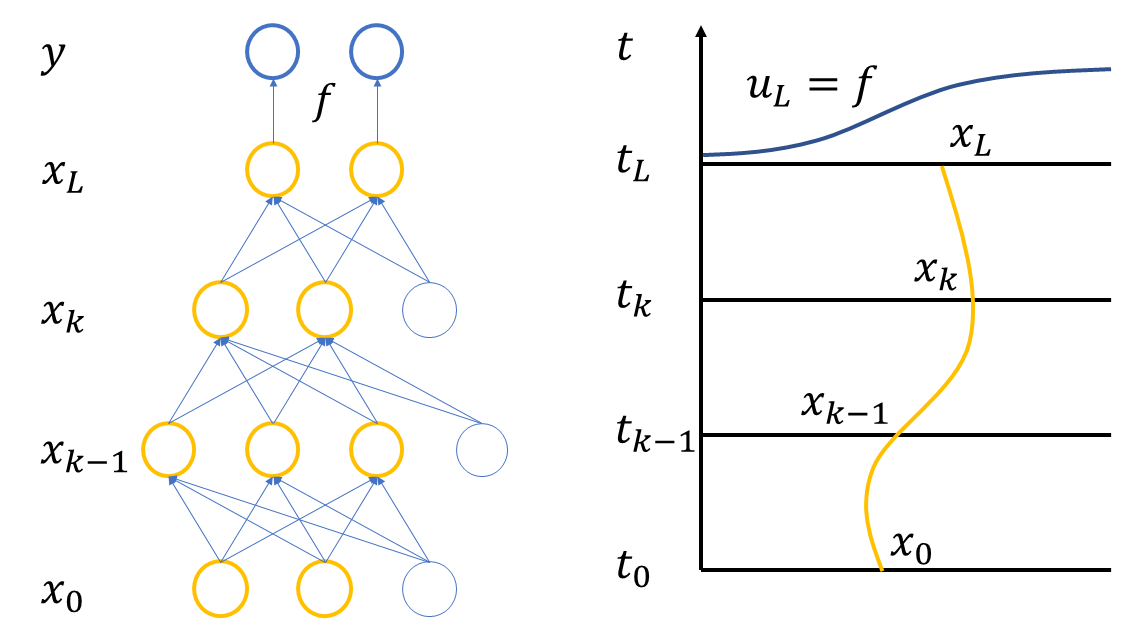}
\caption{Correspondence between neural network and its flow model.}\label{fig_network_transport_correspondence}
\end{figure}

In Section \ref{sec_ResNet} and \ref{sec_plainnet} respectively, we use a transport equation
and its characteristic equation as a continuous flow model for ResNets and plain nets.
This correspondence between neural network and its flow model is very natural, or even obvious for ResNets.
It is summarized in Table \ref{tab_correspondence_linear} 
and illustrated in Figure \ref{fig_network_transport_correspondence}. 
\begin{table}[H]
\caption{Correspondence between neural network and its flow model}\label{tab_correspondence_linear}
\vspace{5mm}
\centering
\begin{tabu} to 1 \textwidth{ c | c }
\hline
\textbf{Neural Network} & \textbf{Flow Model}\\
\hline
layer $k$ & time $t_{k}$\\
\hline
parameters and activations & transport velocity field $v(t, x)$\\
\hline
output function $f$ & terminal value function $u(T, \cdot) = f$\\
\hline
prediction map $F$ & initial value map $u(0, \cdot)$\\
\hline
label $y$ & initial value $u(0, X) = F(X)$\\
\hline
feedforward & solving IVP of characteristic equation\\
\hline
prediction & solving TVP of transport equation\\
\hline
supervised learning & solving inverse problem\\
\hline
\end{tabu}
\end{table}

Inspired by the connection between neural networks and transport equations, 
the well studied methods in the area of differential equations might help to understand 
neural networks or to solve related problems. Here are just a few examples:
\begin{enumerate}
\item We have seen the reason for using 2-layer blocks in ResNet.
In the language of transport equation, the inner parameters are used to specify location
in the space of data. It tells the network where to assign a velocity vector.
The outer parameters are used to adjust the magnitude and direction of the velocity vector
at the specified location. The outer parameters are necessary because ReLU is asymmetric.

\item The correspondence provides one way to see why deep is good for neural networks.
In the perspective of TVP of transport equation, in order to transform the terminal value function to
the initial value function, the transport velocity field needs to be complicated.
To make the discretization converge and to control error, it is necessary to use 
small time step size and many iterations. It allows the discretization to be more regular, 
such that each step makes only a small progress. For neural networks, the transformation 
provided by each layer is also very limited. So it needs more layers to accomplish 
the required deformation of datasets. 

\item In practice, ResNets can usually be significantly deeper than plain nets.
Considering their connections with flow model, the reason for this is quite transparent.
On the one hand, the plain net is equivalent to its flow model, which is constructed in 
Section \ref{sec_plainnet}. On the other hand, the flow model can be discretized
in an iterative way to get a ResNet, as described in Section \ref{sec_rediscretization}.
Combining the two facts, we can say that the ResNet is a refinement of the original plain net. 
Naturally, it is deeper than the plain net. 

\item Although ResNets can be very deep, many authors have shown that the training of ResNets 
is easier than plain nets of comparable depth. From the differential equation point of view,
this is because ResNets deform dataset in an incremental way, which is much regular
than plain nets.

\item When solving PDEs, people often use dissipative terms to increase regularity of solutions. 
In terms of neural networks, it means to add randomness to the feedforward process. 
This idea is very close to the dropout technique \citep{Srivastava2014dropout}.

\item We have already known that ResNet corresponds to method of characteristics for 
transport equations. But there are other methods to solve PDEs \citep{Zhen_ResNetcontrol}, 
which might lead to alternative equivalent architectures to neural networks. 

\item The training of neural networks could be considered as solving inverse problem of
transport equation. It means that both initial value and terminal value are given. 
The task is to find a time-dependent velocity field that transports the initial value 
to the terminal value. Of course, the solution to the inverse problem
is highly non-unique. There are uncountably many velocity fields that can do the job.
Thus the inverse problem is usually formulated as an optimization problem 
constrained by the transport equation as well as the initial and terminal conditions.
There are many methods to solve these problems. Some of them could be modified to train
neural networks \citep{EMSA}.
\end{enumerate}

One possible question about continuous model is the dimension matching problem.
In practice, one has more flexibility to choose different dimensions for different layers.
But in the continuous model, it seems difficult to do so.
Since the main concern of this paper is theoretical, it's not a serious problem for us.
Actually, the dimension matching problem already exists in ResNets.
There is a restriction that the shortcuts are only used when dimensions are matched.
Otherwise, extra projection matrices are needed.
In this note, we have adopted a simple assumption that the dataset is embedded
into a space with sufficiently high dimension at the beginning. This ambient dimension 
doesn't change with time. In order to approximate necessary reduction of intrinsic
dimension of dataset during time, we used compressing flows.
Of course, this theoretical approach is inefficient in practice.
An alternative approach is to glue up different flow models with different dimensions.

\section*{Acknowledgement}
Zhen Li would like to show his gratitude to the support of professors Yuan Yao and Yang Wang 
from the Department of Mathematics, HKUST.

\bibliographystyle{apa}
\bibliography{reference}
\end{document}